# On Triangulating Dynamic Graphical Models

Jeff A. Bilmes　and　Chris Bartels
Department of Electrical Engineering
University of Washington
Seattle, WA 98195

## Abstract

This paper introduces improved methodology to triangulate dynamic graphical models and dynamic Bayesian networks (DBNs). In this approach, a standard DBN template can be modified so the repeating and unrolled graph section may dissect the original DBN time slice and may also span (and partially intersect) many such slices. We introduce the notion of a "boundary" which divides a graph into multi-slice partitions each of which has an interface, and define the "boundary algorithm", a method to find the best boundary (and corresponding interface) between partitions in such models. We prove that, after using this algorithm, the sizes of the best forward- and backward- interface (and also the corresponding fill-ins) are identical. The boundary algorithm allows for constrained elimination orders (and therefore graph triangulations) that are impossible using standard slice-by-slice constrained elimination. We describe the above using the Graphical Model ToolKit (GMTK)'s notion of dynamic graphical model, slightly generalizing standard DBN templates. We report triangulation results on hand-concocted graphs, novel speech recognition DBNs, and random graphs, and find that the boundary algorithm can significantly improve both tree width and graph weight.

## 1 Introduction

Finding high quality Bayesian network triangulations is essential for tractable exact probabilistic inference. Unfortunately, finding the optimum triangulation is NP-complete, so heuristic approaches must be used. A good triangulation, however, is often re-used many times for exact inference thereby amortizing the (sometimes very large) cost of the original triangulation procedure.

Triangulation in dynamic Bayesian networks (DBNs) [5] is distinctively difficult. First, the typical dynamic model is **much** wider than it is taller, rendering standard triangulation heuristics less effective on such graphs. Standard triangulation heuristics include greedy schemes where elimination orders are produced by choosing next nodes according to their current fill-ins, sizes, or weights [15]. These schemes, however, can easily start eliminating nodes with neighbors that span many time slices and thereby produce correspondingly large cliques. Second, evidence will typically come in at different lengths meaning the graphs will vary in size. For example, in speech recognition, evidence corresponds to unknown speech utterances whose time-length will vary between utterances. Therefore, for each evidence set, the graph and the number of random variables will change. The standard approach to triangulation (triangulate the graph once and then reuse it multiple times as evidence comes in) will not work when applied in its simplest of forms — a triangulation for a length $T$ graph might not easily apply to a length $T + 1$ graph.

There are several possible solutions. First, one can re-triangulate the freshly unrolled graph each time evidence becomes available. Second, one can triangulate once using a very long utterance (say length $T$), and hope to find periodicity. One can splice out sections of the triangulated graph (and its corresponding junction tree) and re-form an appropriately triangulated graph for any utterance of length less than $T$. Neither of these approaches is entirely satisfactory however. The first is lacking because finding an optimum triangulation is intractable and running even a poor heuristic multiple times for each length can be wasteful. The second approach is inadequate because given an arbitrary triangulation, one might not find such periodicity. And even if it exists, algorithms for finding it and manipulating the graph can be complex. Both approaches suffer from the fact that as the graph length grows so does the number of possible triangulations thereby making it more difficult (and less likely) to find high quality triangulations. Of course, one can resort to approximate inference techniques in DBNs [6, 12] but with a good triangulation, some even quite complex networks can be utilized exactly.

The most promising work on DBN triangulation and exact inference uses a constrained elimination scheme [8, 10, 16, 4, 12, 13]. In this case, rather than considering all possible elimination orders in an unrolled graph, one places *a priori* constraints on the elimination order that severely restrict the number of elimination orders but (hopefully) do not severely restrict the triangulation quality. Specifically, in slice-by-slice elimination [8, 10, 16, 4, 12] the nodes in slice $t$ are completely eliminated before any nodes in slide $t+1$, making the maximum clique size (roughly) bounded by the "height" of the network. Moreover, rather than unrolling and then triangulating the graph anew for each evidence set, these approaches create fractional slices (a slice plus either its interface to the next slice [8, 10] or its interface to the previous slice [16, 4], and called the "1.5DBN" in [12, 13]). The fractional slices can be triangulated individually, repeated to any desired length, and then stitched together to form a valid unrolled and triangulated graph. Experimental evidence has even shown certain constrained triangulation heuristics to be superior to unconstrained heuristics [4], presumably because the search space is much larger in the unconstrained case.

In this paper, we introduce improved methodology to triangulate dynamic graphical models where a standard DBN template can be modified so that the repeating and unrolled graph section may dissect the original DBN time slice and may also span (and partially intersect) many such slices. We introduce the notion of a "boundary" which divides a graph into multi-slice "partitions" each of which has an interface, and define the "boundary algorithm", a method to find the best boundary (and corresponding interface) between partitions in such models. We also define a "partition algorithm" that utilizes the result. These algorithms operate entirely in the space of undirected rather than directed graphs (meaning a DBN must first be moralized). This significantly simplifies the partitioning step and the interface definitions. We prove that, after using this algorithm, the sizes of the best forward- and backward- interface (and also the corresponding fill-ins) are identical. The boundary algorithm allows for constrained elimination orders (and therefore graph triangula-tions) that are impossible using standard slice-by-slice constrained elimination. We describe and implement the above using the Graphical Model ToolKit (GMTK)'s notion of dynamic graphical model, generalizing on standard DBN templates. We report triangulation results on hand-concocted graphs, novel speech recognition DBNs, and random graphs. Using various quality measures (maxclique size, state-space, etc.), the boundary algorithm can significantly improve both tree-width and graph weight.

Section 2 provides general background on constrainedly triangulated dynamic graphs. Section 3 introduces the GMTK DBN model, one that slightly extends standard DBNs. Section 4 describes the boundary algorithm, and proves that the best left- and right-interfaces are equal in quality. Section 5 describes the new GMTK triangulation engine. Section 6 describes our results, and Section 7 concludes. Throughout this paper, we assume basic knowledge of graphical models [11] and their set-theoretic description.

## 2 Technical Background

A DBN [5] of length $T$ is a directed acyclic graph $\mathcal{G} = (V, E) = (\bigcup_{t=1}^{T} V_t, E_T \cup \bigcup_{t=1}^{T-1} E_t \cup E_t^{\rightarrow})$ with node set $V$ and edge set $E$ comprising pairs of nodes. If $uv \in E$ for $u, v \in V$, then $uv$ is an edge of $\mathcal{G}$. The sets $V_t$ are the nodes at slice $t$, $E_t$ are the intra-slice edges between nodes in $V_t$, and $E_t^{\rightarrow}$ are the inter-slice edges between nodes in $V_t$ and $V_{t+1}$. An undirected *dynamic graphical model* takes a similar form but all edges are undirected. A DBN does not typically have this much flexibility — that is, a DBN is specified using a "rolled up" template giving the nodes that are repeated in each slice, the intra-slice edges among those nodes, and the inter-slice edges between nodes of adjacent slices. This template is then unrolled to any desired length $T$ to yield the DBN $\mathcal{G}$.

The following theorem is relied upon by most work on DBN triangulation:

**Theorem 2.1. Rose (Lemma 4 in [14]).**
*Let $\mathcal{G} = (V, E)$ be an undirected graph with a given elimination ordering that maps $\mathcal{G}$ to $\mathcal{G}' = (V, E')$ where $E' = E \cup F$, and where $F$ are the fill-in edges added during elimination. Then $uv \in E'$ is an edge in $\mathcal{G}'$ iff there is a path with endpoints $u$ and $v$, and where all nodes on the path other than $u$ and $v$ are eliminated before $u$ and $v$.*

This theorem is critical for constrained slice-by-slice DBN elimination schemes for the following reason. If there is a path between two nodes $u, v \in V_t$ where all the path nodes (except the endpoints) lie entirely in previous time slices ($< t$), and if all nodes are eliminated in slices less than $t$ before any in slice $t$, then $u$ and $v$ will be connected after triangulation.

When all nodes earlier than time $t+1$ are eliminated and when there is one connected component per slice, there will be a set of nodes that are forced to be complete in slice $t+1$, namely those nodes entirely in slice $t+1$ that either have parents in slice $t$ or have children with other parents in slice $t$. In a directed model, those nodes have been called the *interface* [8, 10, 4], *backward interface* [16], or the *incoming interface* [13] and have been denoted by $I_t^{\leftarrow} \subseteq V_{t+1}$. Given the sets $V_t^{\leftarrow} \triangleq V_t \cup I_t^{\leftarrow}$, and the "1.5 slice" induced subgraphs $\mathcal{G}_t^{\leftarrow} \triangleq \mathcal{G}[V_t^{\leftarrow}]$, it is possible to form a slice-by-slice constrained elimination by first moralizing $\mathcal{G}_t^{\leftarrow}$ to yield $\mathcal{G}_t^{\leftarrow m}$, next completing the nodes $I_{t-1}^{\leftarrow}$ within $\mathcal{G}_t^{\leftarrow m}$ (since by Theorem 2.1 they would be made complete by eliminating nodes up to slice $t-1$), and finally eliminating all of the nodes $V_t$ within $\mathcal{G}_t^{\leftarrow m}$ to yield the complete set $I_t^{\leftarrow}$. The resulting triangulated 1.5 slice subgraphs can be denoted $\mathcal{G}_t^{\triangleleft} = (V_t^{\leftarrow}, E_t^{\triangleleft})$, where $E_t^{\triangleleft}$ consists of original edges of $\mathcal{G}_t^{\leftarrow m}$ plus the fill-in edges added during elimination. It can be shown (corollary 1 of [10]) that the cliques of these subgraphs form an edge clique cover of a constrainedly triangulated graph $\mathcal{G}^{\triangleleft} = (\bigcup_{t=1}^{T} V_t, \bigcup_{t=1}^{T} E_t^{\triangleleft})$. In particular, the resulting triangulation of $\mathcal{G}$ is one that can be obtained using a constrained slice-by-slice elimination scheme. Therefore, given a DBN template, rather than unrolling to length $T$ and then triangulating the entire graph $\mathcal{G}$, it is possible to triangulate only one instance of the 1.5 slice subgraph (and do it only once), taking the resulting cliques repeated over time as an edge clique cover for a validly triangulated version of $\mathcal{G}$. It is not necessary to re-triangulate the graph for each length $T$, something that can yield large savings.

If, on the other hand, all nodes later than slice $t$ are eliminated before those in slice $t$, certain nodes in slice $t$ will be completed, again by Theorem 2.1. These are the nodes in slice $t$ that have children in slice $t+1$, and have been called the *forward interface* [16, 4] or *outgoing interface* [12] and have been denoted $I_t^{\rightarrow} \subseteq V_t$. One can similarly form 1.5 slice induced subgraphs $\mathcal{G}_t^{\rightarrow} \triangleq \mathcal{G}[V_t^{\rightarrow}]$ where $V_t^{\rightarrow} \triangleq V_t \cup I_{t-1}^{\rightarrow}$, and then moralize $\mathcal{G}_t^{\rightarrow}$, complete the nodes $I_t^{\rightarrow}$, and eliminate nodes $V_t$ yielding the completed $I_{t-1}^{\rightarrow}$ and the corresponding triangulated $\mathcal{G}_t^{\triangleright}$. These subgraphs form an edge clique cover for $\mathcal{G}^{\triangleright} = (\bigcup_{t=1}^{T} V_t, \bigcup_{t=1}^{T} E_t^{\triangleright})$ which is also a triangulation of $\mathcal{G}$ [4, 12], this one yielding a slice-by-slice elimination order in the reverse time direction.

Since moralization can only remove independence properties when going from the directed to the undirected model [11], one can easily see using Markov properties via graph separation in the moralized (and therefore undirected) graphs that either form of interface renders its past conditionally independent of its future. Specifically, we have that $V_{1:t} \perp\!\!\!\perp (V_{t+1:T} \setminus I_t^{\leftarrow}) | I_t^{\leftarrow}$ for the left interface, and $(V_{1:t} \setminus I_t^{\rightarrow}) \perp\!\!\!\perp V_{t+1:T} | I_t^{\rightarrow}$ for the right interface. This is a property any interface must have for it to be useful for inference. Also, since the left or right interfaces are completed in the above two procedures, one can see that a lower bound on the maxclique size of $\mathcal{G}$ using the "better" of the two schemes is $\min(|I_t^{\rightarrow}|, |I_t^{\leftarrow}|)$. Moreover, a naive triangulation of $\mathcal{G}$ would just complete the sets $V_t^{\rightarrow}$ or $V_t^{\leftarrow}$ which corresponds to the worst a slice-by-slice elimination scheme could possibly do [4]. Therefore, an upper bound of the maxclique size of $\mathcal{G}$ using the better of the two schemes, is $\min(|V_t^{\rightarrow}|, |V_t^{\leftarrow}|)$. Therefore, it has been argued that one might choose to use either the left or right interface depending on their sizes. It has been noted that some graphs have smaller interfaces and others have smaller forward interfaces, and that when all "temporal" edges are persistent (between corresponding nodes in successive slices), the left interface can be no better than the right interface [16, 4, 12]. It has also been noted that neither the right nor the left interface need be optimal [16].

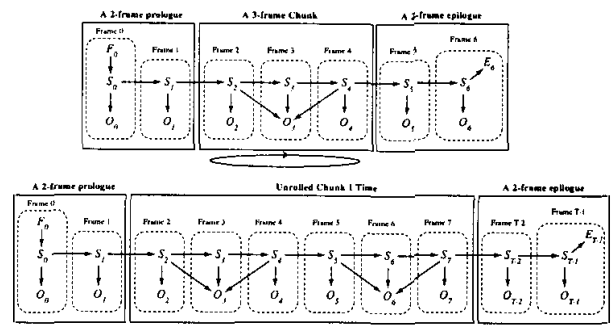

Figure 1: A multi-frame GMTK template (top) with a two-frame prologue $\mathcal{P}$, a 3-frame chunk $\mathcal{C}$, and a 2-frame epilogue $\mathcal{E}$ and unrolled one time (bottom).

## 3 The GMTK Template

Before moving on, we next describe the GMTK template, a generalization of a standard DBN template that also helps to motivate our novel triangulation procedures. Note, however, that the triangulation procedures described in this paper are entirely applicable to standard DBN templates. The graphical modeling toolkit (GMTK) [2] is a general purpose software system for developing graphical-model based speech and language systems. While being graphically oriented, GMTK also has features that are contained in common speech/language toolkits (e.g., pruning, scaling factors, etc.). In this section, we describe only its extended DBN representation — other features are described in [2, 1].

As mentioned above, a typical DBN template is de-scribed using slice nodes and their intra- and inter-set of edges. A GMTK template extends a standard DBN template in three ways: first, it allows for backward time links (so the future need not be independent of the past given the present); second, it allows for slices to span multiple time points, so slices are called *chunks*; third, it allows for a different special structure to occur at both the beginning and the end of the unrolled network. Specifically, a GMTK template consists of a prologue subgraph $\mathcal{P} = (V^p, E^p)$, a chunk subgraph (to be unrolled) $\mathcal{C} = (V^c, E^c)$, an epilogue subgraph $\mathcal{E} = (V^e, E^e)$, and interface edges $E^{pc}$ between $\mathcal{P}$ and $C$, $E^{ce}$ between $\mathcal{C}$ and $E$, and inter-chunk edges $E^{cc}$ between nodes in the previous and current chunk. Each of these subgraphs can be any number of time slices long and we let $T(\mathcal{P})$ denote the number of slices contained within $\mathcal{P}$ (similarly for $\mathcal{C}$ and $\mathcal{E}$). Therefore, the number of slices in an unrolled GMTK-DBN $\mathcal{G}^T$ is allowed to be $T = T(\mathcal{P}) + kT(\mathcal{C}) + T(\mathcal{E})$ for $k$ a positive integer. $\mathcal{G}^T$ may be specified as follows:

$$\mathcal{G}^T = \left(V^p \cup \bigcup_{t=1}^{k} V_t^c \cup V^e, E^{pcc} \cup \bigcup_{t=2}^{k}(E_t^c \cup E_t^{cc}) \cup E^{cee}\right)$$

corresponding to a graph unrolled $k - 1$ times, where $E^{pcc} = E^p \cup E^{pc} \cup E_1^c$ and $E^{cee} = E^{ce} \cup E^e$. Specifying the graph with $k = 1$ corresponds to the *basic GMTK template* $\mathcal{G} = [\mathcal{P}, \mathcal{C}, \mathcal{E}]$, and we refer to $\mathcal{P}$, $\mathcal{C}$, and $\mathcal{E}$ as the template *partitions*.

As mentioned above, the latest GMTK allows not only forward but also backward temporal edges, thereby increasing the size of the family of expressible models (of course, directed cycles are still disallowed). This allows the representation of certain reverse-time causal effects such as coarticulation in human speech, usually defined as a change in the acoustic-phonetic content of a speech segment due to anticipation and/or preservation of adjacent segments — the realization of a segment can thus depend on both the past and the future (see Figure 1).

Note that either $\mathcal{P}$ or $\mathcal{E}$ (but not both) may be empty. Therefore, a GMTK template generalizes and can easily represent a standard DBN — make $\mathcal{E}$ empty, and have $\mathcal{P}$ and $\mathcal{C}$ both be one slice long. We refer to $E^{cc}$ as the basic *boundary edges*.

It is relatively easy to apply the constrained elimination schemes of Section 2 to a GMTK-DBN, but the definition of "interface" must change due to the potential presence of backward time edges. Given a basic GMTK template, let $V_R^p \subseteq V^p$ be the nodes of $V^p$ that either: 1) have children within $V^c$ (corresponding to forward time edges); or 2) have parents within $V^c$ (backward time edges) or have children within $V^p$ having parents in $V^c$ (edges due to moralization).

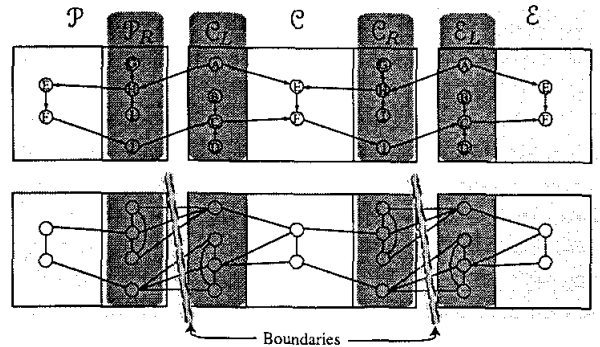

Figure 2: A GMTK-DBN template (top) and its moralization (bottom). Here, $T(\mathcal{P}) = 2$, $T(\mathcal{C}) = 3$, and $T(\mathcal{E}) = 2$. The left and right interfaces for each partition have been labeled in each graph, and is much simpler to define in the undirected version. The basic boundary between subgraphs $\mathcal{P}_R$ and $\mathcal{C}_L$, and the one between $C_R$ and $\mathcal{E}_L$, are both shown as a think line.

Similarly, let $V_L^c \subseteq V^c$ be the nodes of $V^c$ that either: 1) have children within $V^p$ (backward time edges) or 2) have parents within $V^p$ (forward time edges) or have children within $V^c$ having parents in $V^p$ (moralization). One can analogously define $V_R^c$ and $V_L^e$, and the corresponding induced *interface subgraphs* $\mathcal{P}_R$, $\mathcal{C}_L$, $\mathcal{C}_R$, and $\mathcal{E}_L$ as shown in Figure 2.

Rather than continuing to define the interface subgraphs in this convoluted way, it is much simpler to define an interface **after** moralization has taken place and using the resulting undirected graph. We therefore define the *left interface* of a partition to be all nodes that directly connect to the adjacent partition on the left, where adjacency is with respect to the moralized and therefore undirected graph. In Figure 2, the left interface of $\mathcal{C}$ consists of the nodes $\mathcal{C}_L = \{A_3, B_3, C_3, D_3\}$ where 3 is the frame number. As can be seen, it is much easier to determine using the moralized (bottom) graph in Figure 2. We similarly define the *right interface* of a partition to be all nodes that directly connect to the adjacent partition on the right. In other words, constructs such as nodes that have "children within $V^c$ having parents in $V^p$" in this section and in Section 2 is accounted for entirely by the moralization step. Moreover, using the Markov properties of undirected graphs and their correspondence to simple graph separation [11], it is easy to see that the interfaces under these definitions render the left portion of the graph conditionally independent of the right portion (similar to as described in Section 2).

Henceforth, we refer to left and right interfaces using only the undirected dynamic graphical model (one possibly obtained via moralization). In a graph with forward only temporal edges, the left interface will tend to be bigger since moralization will only increase its size. Similarly, a graph having backwards only time edges will tend to have a larger right interface.

A slice-by-slice elimination order therefore applies in the analogous way, but in this context it would be called a *chunk-by-chunk* elimination. For example, with a left interface, one creates a "1.5 chunk" left interface subgraph, say $C_t^x \triangleq C_t \cup C_{(t+1)L}$ (where now $t$ is chunk number), completes the nodes $V(C_{tL})$ and $V(C_{(t+1)L})$ within $C_t^x$, and then eliminates the result to obtain triangulated graph $C_t^\triangleleft$. The analogous result exists for the 1.5 chunk right interface subgraph $C_t^x$. In either case, the boundary (see bottom Figure 2) therefore connects the left interface (i.e., the nodes just on the *right* of the boundary) with the right interface (the nodes just on the *left* of the boundary).

### 3.1 Example GMTK Templates

It is illustrative at this point to examine several GMTK-DBN templates, some of which are currently being used as speech and language research systems. Due to space limitations, details are left unspecified and only graph structures are given (e.g., certain dependencies might be deterministic). The graphs are displayed in Figure 3.[1] The top left (A) shows a standard GMTK template used for a number of speech recognition systems [2, 3]. The top right (B) shows a template currently being used for connected-word continuous speech recognition. The bottom left (C) is a graph used to illustrate a property of the boundary algorithm below. The middle right (D) shows a graph [13] and its 2×-unrolled version where standard slice-by-slice elimination fails to achieve the obvious size-2 maxclique. The bottom right (E) shows a "snake-like" graph, one where no constrained elimination scheme will achieve its size-2 maxclique.

## 4 Boundary Based Triangulation

As can be seen in Figure 2, the basic boundary yields a left and right interface both with size four, implying that using this boundary would produce triangulations with a maxclique of at least that size. The chunk-based view of a frame makes it clear, however, that an improved boundary (and corresponding interface) can be found. Inspecting the figure, the nodes $E_3, F_3$ appear to be candidates for a good (either left or right) interface of size only two (see Top Figure 4). These nodes can thus define one side of a new boundary, but choosing them would break $C$ into two pieces, making standard unrolling impossible. Drawing inspiration from software-pipelining al-gorithms, it is possible using this new boundary to recover an unrollable graph by creating a new chunk $C'$ consisting (on its left) of the second portion of $C$ and (on its right) of the first portion of $C$. The new chunk $C'$ is what gets unrolled, and the residual portions of $C$ get absorbed into $P$ (thereby creating $P'$) and $\mathcal{E}$ (creating $\mathcal{E}'$). For Figure 2, this is depicted in the bottom of Figure 4, and is shown more generally in Figure 5-A. The approach is of course applicable to a standard DBN template, since $C$ can be thought of as one long "slice" even if it corresponds to multiple time slices.

---
[1](B) is by Özgür Çetin, Brian Lucena provided the idea for (C), and (D) is by Kevin Murphy [13]

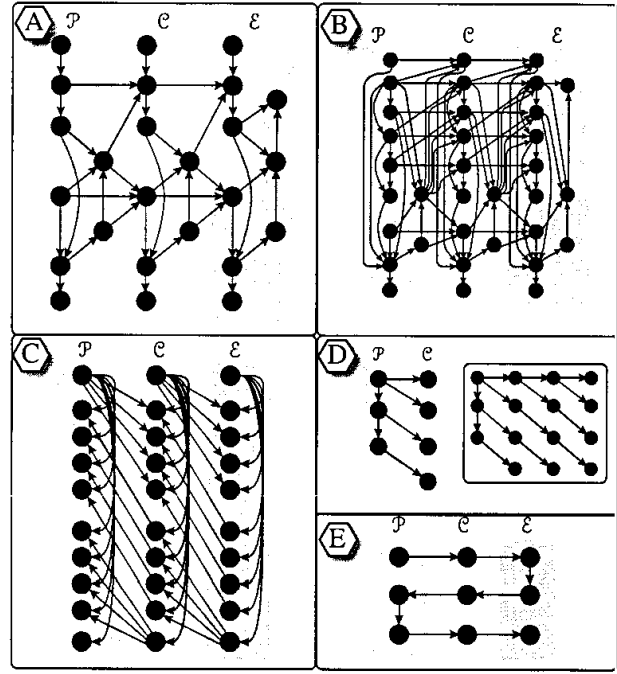

Figure 3: Example GMTK-DBN Templates

More generally still, there is no reason the boundary should be limited only to one chunk — rather, a boundary could instead span across $M \geq 1$ successive chunks. While there is no guarantee that an $(M > 1)$-boundary will yield a better triangulation than $M = 1$ for all graphs, there are indeed certain graphs for which only $M > 1$ will allow a constrained elimination procedure to be optimal. For examples, consider Figure 3-D, where in the unrolled version the maxclique size is two, but a constrained slice-by-slice elimination scheme will produce a maxclique size of at least three (the right interface size). The chunk in this graph, however, is not a connected component. Figure 3-C shows an example where each chunk is indeed connected, but a slice-by-slice elimination will still produce a larger maxclique than necessary (the tree width of this graph is only 4, also see Figure 7 for a similar example). These graphs demonstrate that if a boundary is allowed to span multiple chunks (or slices in a standard DBN), it may be possible to ob-tain better triangulations. In Figures 3-C and -D, the boundary would need to span 3 chunks.

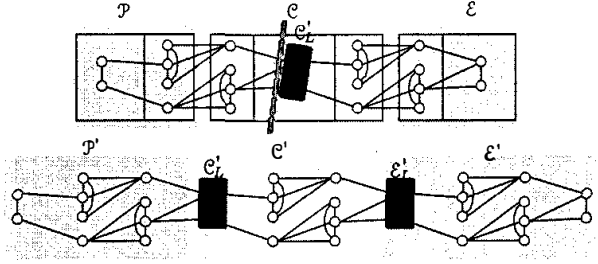

Figure 4: Top: Figure 2 but with a different boundary, one that has only two nodes adjacent on its right. Bottom: a re-partitioning of the graph based on this new boundary. This re-partition defined a new and better GMTK-DBN template $\mathcal{G}' = [\mathcal{P}', \mathcal{C}', \mathcal{E}']$.

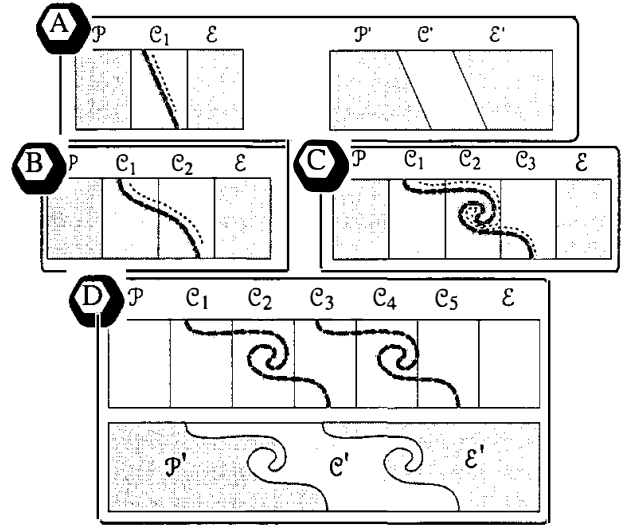

Figure 5: Boundary example for $M = 1$ (A-left) and its repartitioning with $S = 1$ (A-right), and boundary example for $M = 2$ (B) and $M = 3$ (C). Left interfaces are shown as dashed lines. D: Graph repartitioning with a new boundary and $M = 3$ and $S = 2$.

We now define the *boundary algorithm*, a method to find the optimal chunk-spanning boundary. Given partitions $\mathcal{P}$, $\mathcal{C}$, and $\mathcal{E}$, define $\mathcal{C}^M \triangleq \bigcup_{t=1}^{M} \mathcal{C}_t$ as $M$ copies of chunk $\mathcal{C}$ corresponding to the GMTK template unrolled $M - 1$ times. If $\mathcal{P}$ or $\mathcal{E}$ is empty, we unroll one additional time and replace the missing partition with an additional single copy of $\mathcal{C}$. For simplicity, the algorithm will be described using set operators on graph names, but they actually operate on the graphs' vertex sets. Also, define $J()$ to be a function on left interfaces that provides a numerical rating of the interface quality (discussed further in Section 4.2). The boundary algorithm is defined as follows:

1: **Function** Boundary($\mathcal{P}, \mathcal{C}^M, \mathcal{E}$)
2: Let $\mathcal{C}_L$ be the left interface of $C^M$.
3: Note current interface & quality $J(\mathcal{C}_L)$.
4: Call **BoundaryRecurse**($\mathcal{C}_L, \emptyset$).
5: **Function** BoundaryRecurse($\mathcal{C}_L, \mathcal{B}_L$)
6:    **for all** $v \in \mathcal{C}_L$ **do**
7:       **if** $(ne(v) \cap \mathcal{E}) \neq \emptyset$, **continue**.
8:       $\hat{\mathcal{B}}_L \leftarrow \mathcal{B}_L \cup \{v\}$.
9:       **if** $\hat{\mathcal{B}}_L$ contains entire first chunk, **continue**.
10:      $\hat{\mathcal{C}}_L \leftarrow \left(\mathcal{C}_L \cup (ne(v) \cap \mathcal{C}^M)\right) \setminus \hat{\mathcal{B}}_L$.
11:      **if** memoized($\hat{\mathcal{C}}_L$), **continue**.
12:      Note current interface & quality $J(\hat{\mathcal{C}}_L)$.
13:      Call **BoundaryRecurse**($\hat{\mathcal{C}}_L, \hat{\mathcal{B}}_L$).
14:    **end for**

The algorithm starts out with the standard left interface, and at each step advances the boundary across a single node in the current interface (see Figure 6). At each boundary advance, the algorithm defines a new boundary and a new corresponding left interface. The algorithm considers all possible left interfaces — this is true since given any left interface, there is a path of reverse boundary advances that will lead to the initial left interface. The algorithm uses an auxiliary variable $\mathcal{B}_L$, consisting of the nodes past which the boundary has advanced at a given moment — in other words, one might say that these are nodes to the "left" of the current boundary but that lie entirely within $\mathcal{C}^M$ (thus, $\mathcal{B}_L$ starts out empty, line 4). Note, the left interface consists of the nodes that are directly to the "right" of the boundary and that have nodes adjacent to the boundary. The routine (line 5) goes through each element $v$ (line 6) in the current left interface and advances the boundary past that node. Given this new boundary, it creates a new left of boundary set $\hat{\mathcal{B}}_L$ (line 8). The check (line 7) ensures termination by not letting a boundary advance too far — in particular, the boundary never advances beyond the point where its left interface is identical to the basic starting right interface. An additional check (line 9) ensures that a boundary does not move entirely beyond an entire chunk, since that would lead to redundant boundaries. The new left interface $\hat{\mathcal{C}}_L$ (line 10) is constructed starting with the old left interface $\mathcal{C}_L$, adding the neighbors of $v$ that are on the right of the new boundary (all neighbors of $v$ are initially added, but those in $\hat{\mathcal{B}}_L$ are removed), and ensuring $v$ is not part of the new left interface (subtracting off $\hat{\mathcal{B}}_L$ removes $v$). Since the same boundary could be encountered multiple times, a memoizing check ensures that this does not happen (line 11). If the interface quality is better than what has been seen so far, the current in-terface and its quality is remembered (line 12). Lastly, the algorithm is called recursively on the new interface $\hat{\mathcal{C}}_L$ and the left of boundary set $\hat{\mathcal{B}}_L$. An example of a run of this algorithm is given in Figure 6, and Figure 7 shows how the set of nodes comprising the optimal interface can sometimes be quite unexpected.

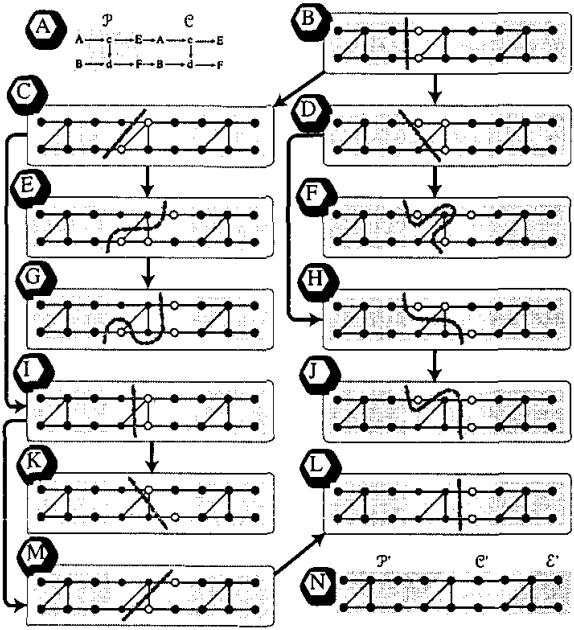

Figure 6: Example of the boundary algorithm to find the best left interface. Thick dark arrows depict the tree structure of the recursive algorithm, and boundaries and their interfaces are shown for each call. **A:** a basic GMTK template with $\mathcal{E} = \emptyset$. Capital letter named variables have high cardinality, and lower-case letter variables have low cardinality. **B:** The starting point of the algorithm. The initial boundary is shown as a thick line and the basic left interface is indicated by white nodes. **C:** the boundary advances past node $A$, leading to a new left interface (white nodes) and left of boundary set $\mathcal{B}_L$ (red/gray nodes). **D:** the boundary advances past node $B$. **E-M:** the various boundaries (as think curves), left interfaces (as white nodes), and left of boundary sets $\mathcal{B}_L$ (red/gray nodes) are shown, where D calls F and H, C calls E and I, and so on. **N:** the partitions defining the new template $G'$ corresponding to boundary C.

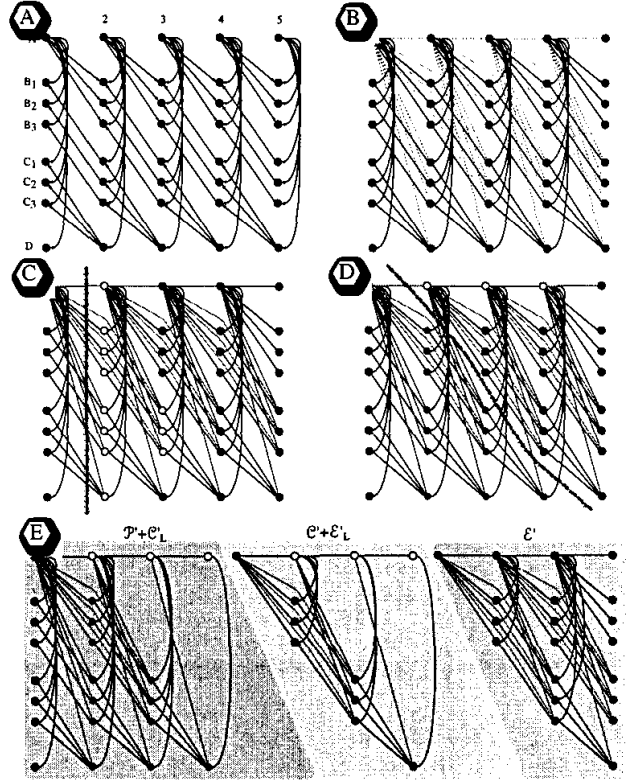

Figure 7: An optimal interface can sometimes be surprising. **A:** a graph similar to Figure 3-C. On a first glance, it appears the best interface size in this graph could be quite large; **B:** moralized version, where additional edges are dashed; **C:** the initial boundary and left interface (white nodes); **D:** an optimal boundary, left interface (white nodes), and $\mathcal{B}_L$ (red/gray nodes). Therefore, in this graph the set of nodes $\{A_t, A_{t+1}, A_{t+2}\}$ separates the graph into two conditionally independent subgraphs. **E:** The resulting template partitions *with* interfaces using boundary D.

Interestingly, a right interface version of the algorithm is obtained simply by invoking the procedure using Boundary($\mathcal{E}, \mathcal{C}^M, \mathcal{P}$) (i.e., swapping the first and third argument). In such case, the initial left interface becomes the standard right interface, and the boundary advances from right to left across nodes. Note further that the boundary algorithm defines the optimal boundary implicitly via the optimal (left or right) interface that it produces. To get the actual boundary edges $E^{cc}$, we simply take the left (resp. right) adjacent edges of the left (resp. right) interface.

All interfaces considered by the boundary algorithm will render the left of the graph conditionally independent of the right portion given the interface. Clearly this is true for the initial interface. Given an interface having this property, a new interface is formed by moving the boundary over one interface node $v$. But separation is preserved since all neighbors of $v$ on the right of the new boundary are added to the new left interface. Therefore, we have the following theorem:

**Theorem 4.1. Separation Property of Interface**
*The boundary algorithm only considers boundaries and their interfaces that separate the left from the right of the graph, and therefore make the left conditionally independent of the right given the interface.*

A small amount of analysis makes it quite clear that the algorithm has (at least) exponential complexity. When considering the simple horizontal ladder graph, for example, the complexity grows as $O(3^n)$ where $n$ is the number of slices in the chunk. Therefore, when considered together with the triangulation problem, we arrive at an exponential number of separate NP-complete problems. Fortunately, most of the graphs we have encountered are small enough that one can run the complete boundary algorithm in at most from a few seconds to about an hour's worth of wall-clock time on a modern workstation. More importantly, once a good boundary (and corresponding triangulation) has been discovered, the cost is amortized over the usage life of the graph (which can be years). This makes it well worth any initial effort in finding a good boundary and good triangulation. In any case, Section 7 discusses future plans for a greedy approximate boundary algorithm.

### 4.1 Boundary-Dependent Graph Re-Partitioning

Given a boundary, one must next use it to partition the graph $\mathcal{G}$. In our approach, we start with a standard template $(\mathcal{P}, \mathcal{C}, \mathcal{E})$, partially unroll it to obtain $\mathcal{G}'$, and then use the boundary to re-partition the partially unrolled graph giving new partitions $[\mathcal{P}', \mathcal{C}', \mathcal{E}']$. The new graph and partition is treated as an original, and unrolling becomes a matter of repeating $\mathcal{C}'$.

Given boundary edges $E^{cc}$ spanning $M$ chunks, one still has an option regarding how many chunks to skip between each boundary. We call this the *chunk skip* parameter $S \geq 1$. Given a GMTK-DBN template, the approach is to partially unroll it $M + S - 1$ times thus allowing room enough for two boundary edge sets $E_1^{cc}$ and $E_2^{cc}$ spaced $S$ chunks apart. The first boundary is "layed across" the first $M$ chunks, and the second boundary is layed across chunks $S + 1$ through $S + M$. These boundaries then re-partition the graph into the new graph $\mathcal{G}^{(M,S)} = [\mathcal{P}', \mathcal{C}', \mathcal{E}']$. This is depicted in Figure 5-D for $M = 3$ and $S = 2$. We thus have the *partition algorithm*.

1: **Function** Partition$(\mathcal{G}, M, S)$
2: From $\mathcal{G} = [\mathcal{P}, \mathcal{C}, \mathcal{E}]$, unroll to extract $\mathcal{C}^M$.
3: Call **Boundary**$(\mathcal{P}, \mathcal{G}^M, \mathcal{E})$ to obtain $E^{cc}$.
4: From $\mathcal{G}$, unroll $M + S - 1$ times to extract $\mathcal{C}^{M+S}$.
5: Create boundary $E_1^{cc}$ spanning chunks 1 through $M$ and boundary $E_2^{cc}$ spanning chunks $S + 1$ through $S + M$.
6: $\mathcal{P}' \leftarrow \mathcal{P} \cup \text{L-cut}(E_1^{cc}, \mathcal{C}^{1:M})$.
7: $\mathcal{E}' \leftarrow \mathcal{E} \cup \text{R-cut}(E_2^{cc}, \mathcal{C}^{S+1:M+S})$.
8: $\mathcal{C}' \leftarrow \text{R-cut}(E_1^{cc}, \mathcal{C}^{1:M+S}) \cap \text{L-cut}(E_2^{cc}, \mathcal{C}^{1:M+S})$.
9: Return $\mathcal{G}^{(M,S)} = [\mathcal{P}', \mathcal{C}', \mathcal{E}']$.

Note that a boundary $E^{cc}$ cuts a collection of chunks into two pieces, the "left cut" (L-cut) and the "right cut" (R-cut). Therefore, the function L-cut$(E_1^{cc}, \mathcal{C}^{1:M})$ returns the nodes to the left of $E_1^{cc}$ within $\mathcal{C}^{1:M}$, R-cut$(E_1^{cc}, \mathcal{C}^{1:M})$ returns the nodes to the right of $E_1^{cc}$ within $\mathcal{C}^{1:M}$, an so on. Since the boundary can be arbitrarily shaped, a "left cut" means the sub-graph that is connected to nodes on the left-most side of the graph (and analogously for right cut). For example, in Figure 5-D, we have that L-cut$(E_1^{cc}, \mathcal{C}^{1:3}) = \mathcal{P}' \setminus \mathcal{P}$.

With a boundary spanning $M$ and skipping $S$ chunks, the use of a re-partitioned GMTK-DBN template $\mathcal{G}^{(M,S)}$ implies that the number of slices in unrolled graphs must correspond to $T = T(\mathcal{P}) + (M + kS)T(\mathcal{C}) + T(\mathcal{E})$ for $k$ a positive integer.[2]

### 4.2 Measuring Boundary Quality

There are a number of different ways of measuring boundary quality. Three simple ways are the interface size $J(\mathcal{C}_L) = |\mathcal{C}_L|$, the number of fill-in edges (i.e., $J(\mathcal{C}_L) =$ the number of edges needed to complete $\mathcal{C}_L$), and interface weight (the state space of the collection of random variables contained within $\mathcal{C}_L$). In each case, the quality measure is *local*, meaning one never looks outside the interface itself to judge its quality. Interestingly, the quality the best left and best right interface will be identical under these $J()$'s.

**Theorem 4.2. Left & Right Interface Parity**
*When $J()$ is local, running the left-interface algorithm Boundary$(\mathcal{P}, \mathcal{C}^M, \mathcal{E})$ will produce an identical quality interface as when running the right-interface algorithm Boundary$(\mathcal{E}, \mathcal{C}^M, \mathcal{P})$.*

*Proof.* Let $\mathcal{C}_L^*$ be the best left interface. Move the left interface nodes to the left of the interface's boundary. These nodes become a right interface for the new boundary. Since the boundary algorithm searches all boundaries, it will always find the best both left and right interface, which from the above are identical. The other direction clearly holds by symmetry. □

There are measures of interface quality other than the local ones mentioned above. A number of *global* quality measures $J(\mathcal{C}_L)$ for a given interface $\mathcal{C}_L$ are also possible, global since $J()$ is a function of the entire graph. These include: 1) the tree width of the resulting triangulated graph $\mathcal{G}^q$; 2) the tree width of the resulting repeated chunk; 3) the state space of the resulting triangulated graph; or 4) the state-space of the repeated chunk (this last one is particularly important since this indicates the degree to which complexity grows with unrolling amount $k$). Within each of the above lie also the different options for triangulating a graph (heuristics, annealing, etc.). With a global measure, therefore, one is not guaranteed that the left and right interface are identical unless one can solve the optimum triangulation problem. From a heuristic perspective, therefore, one might try both. Fortunately, it is easy using the boundary algorithm to do both as mentioned above.

---
[2]Note that it is also possible to append extra subgraphs at the end in order to allow for any number of slices.

## 5 GMTK Triangulation Search Engine

The algorithms above were recently implemented into the GMTK system along with all aforementioned $J()$ functions. Still, a number of ways exist to triangulate a set of partitions. The GMTK triangulation engine solves this using multiple prioritized heuristics. The heuristics include clique size, fill-in, weight, temporal position, file position, user-supplied hint, and random. The heuristics are provided in order by the user. The highest-priority heuristic is used to determine an elimination order, with lower priority heuristics used only to break ties when they occur. GMTK also supports simulated annealing [9] and maximum cardinality search.

If chunks are small enough, it is possible even to exhaustively search all elimination orders. More interestingly, it is possible to produce an exhaustive search over *all* triangulations (the space of triangulations via elimination do not span the space of all triangulations of a graph). In this latter case, it is possible to produced constrained triangulation schemes that lie outside the space of unconstrained triangulations by elimination, sometimes very useful when deterministic and sparse implementations of dependency exist. GMTK supports both methods of exhaustive search.

Users of GMTK, however, often do not wish to concern themselves with the intricacies of graph triangulation. Therefore, GMTK supports a simple *anytime algorithm* where an amount of time is given (1 minute, 2 hours, 3 days, etc.), and the engine searches for the best triangulation possible in that amount of time. We have found this approach quite satisfying from the toolkit user's perspective — one can provide the time they are willing to spend triangulating (a 3-day weekend) before using the graph for research purposes.

## 6 Triangulation Results

This section provides initial results on hand-concocted graphs, random graphs, and DBNs used in speech recognition research systems.

Table 1 shows results for the hand-concocted graphs from Figures 2, 3, 6, and 7. It also gives results for graphs given in [4] and [12]. The columns give the number of nodes in the resulting interface and largest clique ("mc" for maximum clique) from the triangulation. A graph's *weight* is the log base 10 of its state space. In the case of Figure 6 the graph weight for a network unrolled 500 times is listed because the maximum clique size stays constant. Results are given using the minimum of the basic left and right interfaces ($|I_t^{\rightleftarrows}|$), and using the boundary algorithm with $M = 1, 2, 3$ all with $S = 1$ where the left interface size $|\mathcal{C}_L|$ is reported. As can be seen, both the interface and the clique size can improve dramatically.

Table 2 shows results for randomly generated graphs (using methods based on [7]). The first five graphs contain forward only temporal edges and the second five contain both forward and backward. Each network contains 5, 10, 15, or 20 nodes per frame with random variable cardinalities chosen uniformly at random from 2 to 50. All the weights are given for a network unrolled 500 times. All graphs were first partitioned using the basic left and right interfaces with the smallest size. The sizes of the left and right interfaces are given in the first two columns. The partitions were triangulated using all available methods and the size of the smallest maximum clique is reported. The same partitions were triangulated again optimizing for weight. Next, partitions were created using boundary with $M = S = 1$, with $M = 2, S = 1$, with $M = 1, S = 2$, and with $M = S = 2$. The size of the interface and the best maximum clique size are reported. The graphs were partitioned and triangulated separately optimizing for weight. The boundary algorithm improved clique size in four of the graphs, and improved state space in five. In one case the state space was over 80 times smaller. The results were typically worse using the bulkier partitions with $M$ or $S$ greater than one, but in one case[†] $M = 2, S = 1$ gave the best clique size (also*, $|\mathcal{C}_L| = 10$ corresponds to the maxclique optimization but was = 11 for the weight optimization). In another case[‡] $M = S = 2$ gave the best weight. In all the others, $|\mathcal{C}_L|$ was identical for the two strategies.

Table 3 shows weights for the speech research systems [3]. The first column shows our baseline results using the triangulation method (the Frontier algorithm [18]) used in [2]. The second column is the best weight from partitions created from the standard forward/backward interface with minimum size. The third column is the best weight from a variety of boundary partitions. The boundary algorithm shows improvements in two of the graphs. Although boundary shows a definite advantage, the results are not as dramatic as with the random or concocted graphs. An explanation is that the random graphs have equal probability of an edge between variables within a frame and between variables in adjacent frames. Real world graphs tend to be more densely connected within the frame and have fewer temporal edges.

---

[3]Livescu Decode A & B are by Karen Livescu

Table 1: Results on hand-concocted graphs.

| | ($|I_t^{\leftarrow}|, |I_t^{\rightarrow}|$) | | $M=1$ | | $M=2$ | | $M=3$ | |
|---|---|---|---|---|---|---|---|---|
| | $|I_t^{\rightleftarrows}|$ | mc | $|e_L|$ | mc | $|e_L|$ | mc | $|e_L|$ | mc |
| Figure 2 | 4 | 5 | 2 | 4 | 2 | 4 | 2 | 4 |
| Figure 3-C | 9 | 10 | 9 | 10 | 6 | 7 | 3 | 5 |
| Figure 3-D | 3 | 4 | 3 | 4 | 2 | 3 | 1 | 2 |
| Figure 7 | 7 | 8 | 7 | 8 | 5 | 6 | 3 | 5 |
| Fig 2 of [4] | 1 | 3 | 1 | 3 | 1 | 3 | 1 | 3 |
| Fig 3.14 of [12] | 3 | 4 | 3 | 4 | 2 | 3 | 1 | 3 |
| Figure 6 | 2 | 8.90 | 2 | 8.60 | 2 | 8.60 | 2 | 8.60 |

Table 2: Results on random graphs, 500× unrolling.

| | $min(|I_t^{\leftarrow}|, |I_t^{\rightarrow}|)$ | | | | Boundary | | |
|---|---|---|---|---|---|---|---|
| Nodes | $|I_t^{\leftarrow}|$ | $|I_t^{\rightarrow}|$ | mc | Weight | $|e_L|$ | mc | Weight |
| →5 | 4 | 3 | 5 | 10.2735 | 2 | 5 | 10.2733 |
| →10 | 9 | 9 | 11 | 16.2617 | 8 | 10 | 15.0698‡ |
| →15 | 12 | 13 | 13 | 20.2952 | 10* | 12† | 19.3594 |
| →15 | 13 | 11 | 12 | 16.5115 | 9 | 11 | 14.7054 |
| →20 | 16 | 17 | 17 | 25.4712 | 14 | 17 | 23.5510 |
| ↔5 | 4 | 5 | 6 | 12.0194 | 4 | 6 | 12.0194 |
| ↔10 | 8 | 10 | 9 | 14.9034 | 8 | 9 | 14.9034 |
| ↔15 | 14 | 13 | 14 | 21.0783 | 12 | 13 | 20.4408 |
| ↔15 | 14 | 13 | 14 | 22.1653 | 12 | 14 | 22.1653 |
| ↔20 | 18 | 20 | 19 | 27.0521 | 18 | 19 | 27.0521 |

Table 3: Weights on speech graphs, 500× unrolling.

| Structure | Baseline | $min(|I_t^{\leftarrow}|, |I_t^{\rightarrow}|)$ | Boundary |
|---|---|---|---|
| Figure 3-A | 6.40814 | 5.8020 | 5.6603 |
| Figure 3-B | 14.2418 | 11.7260 | 11.7260 |
| Livescu Decode A | 11.2024 | 10.9910 | 10.9907 |
| Livescu Decode B | 7.03116 | 6.7382 | 6.7382 |
| Muli-Stream [17] | 8.36556 | 7.4553 | 7.3595 |

## 7 Discussion

In this paper, we introduced the boundary algorithm, a new method for facilitating the triangulation of dynamic graphical models. We plan in future work to define and experiment with greedy and randomized approximate boundary procedures. We also plan to develop a better theoretical understanding of the properties of dynamic graphs and their relationship to $M$ and $S$ in order to predict a-priori the best values of $M$ and $S$ to use.

This work greatly benefited from discussions with both Thomas Richardson and Brian Lucena. We also wish to thank the three anonymous reviewers for their useful comments on clarifying the paper. This work was supported by NSF grant IIS-0093430 and an Intel Corporation Grant.